# Exploration of Attention Mechanism-Enhanced Deep Learning Models in the Mining of Medical Textual Data

Lingxi Xiao[1], Muqing Li[2], Yinqiu Feng[3], Meiqi Wang[4], Ziyi Zhu[5], Zexi Chen[6]

[1]Georgia Institue of Technology, USA

[2]University of California San Diego, USA

[3]Columbia University, USA

[4]Brandeis University, USA

[5]New York University, USA

[6]North Carolina State University, USA

*Abstract*—The research explores the utilization of a deep learning model employing an attention mechanism in medical text mining. It targets the challenge of analyzing unstructured text information within medical data. This research seeks to enhance the model's capability to identify essential medical information by incorporating deep learning and attention mechanisms. This paper reviews the basic principles and typical model architecture of attention mechanisms and shows the effectiveness of their application in the tasks of disease prediction, drug side effect monitoring, and entity relationship extraction. Aiming at the particularity of medical texts, an adaptive attention model integrating domain knowledge is proposed, and its ability to understand medical terms and process complex contexts is optimized. The experiment verifies the model's effectiveness in improving task accuracy and robustness, especially when dealing with long text. The future research path of enhancing model interpretation, realizing cross-domain knowledge transfer, and adapting to low-resource scenarios is discussed in the research outlook, which provides a new perspective and method support for intelligent medical information processing and clinical decision assistance. Finally, cross-domain knowledge transfer and adaptation strategies for low-resource scenarios, providing theoretical basis and technical reference for promoting the development of intelligent medical information processing and clinical decision support system.

## I. INTRODUCTION

The explosive growth of information in the medical field and the heterogeneity of data, such as medical records and medical literature, have become major challenges in medical information processing. Traditional text mining methods are weak in the face of this complex and huge data, mainly because of semantic complexity, domain specialization, data sparsity, and other problems restrict its performance. Deep learning models based on attention mechanisms have attracted much attention. The attention mechanism mimics the human reading process, being able to automatically learn key information from text and to weight the information to better understand and represent the text[1].

This paper aims to verify its effectiveness and superiority in medical information processing through comparative experiments. First, we will introduce the principles of attention mechanisms and how they can be applied to deep learning models. Then, we discuss how to apply attention mechanism to medical text mining tasks to improve text understanding and information extraction. Finally, the performance of the attention-mechanism-based deep learning model in medical text classification, named entity recognition, and other tasks is verified by experiments, and its potential application value in future medical information processing is prospected.

This study is committed to an in-depth analysis of the application potential and practical effects of the attention-mechanism-based deep learning model in the field of medical text mining, aiming to comprehensively verify its significant advantages in improving the accuracy and efficiency of medical information processing through rigorous comparative experiments. Firstly, we will explore the basic concept of the attention mechanism, its working principle, and its flexible integration strategy in various deep learning architectures from a theoretical level, and show how it builds a sensitive perception mechanism for text details inside the model.

Subsequently, the paper will elaborate on how the attention mechanism is deeply integrated with the specific tasks of medical text mining, including but not limited to the core fields of medical text classification, named entity recognition, and reveal its unique value in enhancing the model's understanding of medical terms, complex context analysis, and implicit information mining. Through a carefully designed experimental framework, this study will quantitatively evaluate the performance indicators of the attention-mechanism-based model when processing actual medical data sets, and compare and analyze its significant improvement compared with traditional methods.

## II. RELATED WORK

In recent years, medical text mining research has ushered in a new stage of development, which has built a solid theoretical and practical foundation for this research project. This section aims to lay out a clear path for innovative applications of attention-mechanism-based deep learning models in medical information extraction and analysis by comprehensively reviewing and dissecting key recent contributions in several core research areas. A comprehensive review of the work of these pioneers will clarify the current research frontier, provide a valuable benchmark, and further reveal how attention mechanisms can realize their unique value and potential in the highly specialized and complex data environment of medical texts. This discussion will delve into core issues across several areas: the mature application of medical text classification; the innovative role of attention mechanisms in named entity recognition tasks; recent progress in detecting drug side effects and extracting relationships; the rise of pre-trained models in the medical field; and active exploration in privacy protection and data efficiency improvement. These developments have laid a solid theoretical and practical foundation for subsequent research.

Previous research concentrated on employing convolutional neural networks[2] and recurrent neural networks[3] for the classification of medical texts, such as the RNN-based model for disease classification. Subsequently, Xu et al., by combining the BiLSTM [4] and the attention mechanism [5-7], significantly improved the accuracy of the model in identifying disease categories in medical records. This work highlights the potential of attention mechanisms in complex medical entity recognition tasks.

The automatic detection of drug side effects is one of the important applications of medical text mining. Using Transformer architecture [8-10] and multi-head attention mechanism, a model was built that can automatically extract drug side effects from large-scale medical literature, significantly improving the recall rate and accuracy. By introducing the domain knowledge-guided attention mechanism, the model's performance on sparse data was effectively improved, proving the importance of domain knowledge fusion. With the emergence of BERT [11] and its subsequent variants, the application of pre-trained models in medical text processing has become a new hotspot. BioBERT [12], introduced by Alsentzer et al., significantly improves the model's performance in medical text classification and entity recognition tasks through additional pre-training on medical texts, and provides high-quality initialization parameters for deep learning models based on attention mechanisms. This mechanism has not only achieved significant milestones in areas such as computer vision[13-16], medical image[17-20], and semantic mining [21-22], but has also demonstrated considerable potential in medical healthcare, particularly in medical disease association network analysis[23-24]. Its robust ability to learn structural representations offers fresh insights and instruments for medical decision-making.

Given the sensitivity of medical data, some research is beginning to explore how to learn effectively while maintaining privacy. Cui et al. Proposed a medical text mining method under the federal learning framework [25], which can train the model without directly accessing the original data, providing a new idea for processing medical data. At the same time, to solve the problem of sparse medical text data, researchers have also developed lightweight models and transfer learning strategies to reduce the dependence on large amounts of labeled data. On this basis, this study will further explore the deep integration of adaptive attention mechanism and domain knowledge, aiming to provide more advanced and practical solutions for efficient and secure processing of medical information.

## III. THEORETICAL BASIS

### A. Attention mechanism

The attention mechanism first appeared in the field of computer graphics before finding extensive application in natural language processing, speech recognition, and additional tasks. Its design is inspired by the shift of visual focus when humans observe things, that is, the eyes naturally focus on the area of concern, often referred to as the focus of attention. This mechanism automatically captures critical information while filtering out nonsense to improve task performance.

In the field of natural language processing, attention mechanisms were originally applied to Machine Translation tasks (Machine Translation[26], MT). In 2021, Tomar and other scholars proposed for the first time to use attention mechanism as Decoder [27] in this task to automatically extract key information related to target words in source statements and build context vectors, replacing semantic vectors generated by traditional RNN encoders. Since then, more and more scholars have begun to apply attention mechanism in named entity recognition tasks. Its primary principle involves computing the similarity between the current input unit and the entire input statement using a similarity function and assigning the resultant calculation as a weight to the input statement[28]. Typically, additive models, dot product models, scaled dot product models, and other functions serve as similarity functions, as depicted in the equation:

This formula is a functional expression in deep learning, or machine learning, that represents different types of models. To be specific:

Additive model: $f(Q, K_i) = v^T \tanh(WQ + UK_i)$

Dot product model: $f(Q, K_i) = Q^T K_i$

Scaled dot product model: $f(Q, K_i) = \frac{Q^T K_i}{\sqrt{d}}$

Bilinear model: $f(Q, K_i) = Q^T W K_i$

Where $Q$ and $K_i$ are input vectors, and $v$, $W$, $U$, and $d$ are constants or parameters. These models can be used for various tasks, such as text matching, recommendation systems, and more. Additive and bilinear models are generally more complex than dot product models, but they can capture more nonlinear relationships. The scaled dot product model is an improvement on the dot product model, which can prevent numerical overflow problems.

In the preceding process, parameter matrices v, W, and U are parameters that the network needs to learn, where K represents Key and Q represents Query. By using softmax function to normalize the score calculated by similarity, the probability distribution of all elements can be obtained, and these elements have weight coefficients, so that the sum value of all coefficients is 1, thus strengthening the weight of important information. The details are as follows:

$$\alpha_i = \text{softmax}(f(Q, K_i)) = \frac{exp(f(Q, K_i))}{\sum_j exp(f(Q, K_j))} \quad (1)$$

The similarity algorithm calculates the address K in both the Query and memory; based on these calculations, the relevant element V is selected. The aggregation of these values results in the V corresponding to each address K. This process involves determining the similarity between the Query and Key, subsequently deriving weights for each V value. Following this, weights are assigned to each V value. Subsequently, a weighted summation of the V values is executed to yield the final Attention value, as depicted below:

$$\text{Attention}(Q, K, V) = \sum \alpha_i V_i \quad (2)$$

*B. Pre-trained language models*

In the field of NLP, converting text statements to vector form is the first task of each task. Word vector, also known as word embedding, is a way to represent a word or word in a natural language text using a low-dimensional real number vector. Different words or words have different vector representation, which can effectively mine the semantic information contained in the text. The traditional unique heat coding method has high latitude and simple performance, and can not accurately reflect the semantic relationship between words[29].

Because neural networks have excelled in various fields, scholars have begun to use neural networks to train language models to acquire word vectors. The main task of the language model is to calculate the correct text probability P for a given sequence of words. The following formula shows the calculation method of p-value, where if the given word sequence conforms to the grammatical rules, the P-value is higher; otherwise, the P-value is lower, which means that the given word sequence does not conform to the grammatical logic.

$$P(w_1, w_2, \ldots, w_m) = P(w_1) \ldots P(w_m | w_1, w_2, \ldots, w_{m-1}) \quad (3)$$

Training language models does not require highly labeled data, and can be trained on large-scale corpora. The training process is mainly to predict the next word of the current word according to the context of the corpus, and constantly adjust the model parameters through random initialization and backpropagation algorithm. Instead of the traditional random initialization method, the pre-trained language model initializes its parameters with those trained on a large-scale dataset beforehand. Then train or fine-tune for specific tasks to get good results. Currently, the most popular Pre-Training language models include GPT (Generative Pre-training), BERT and XLNet.

GPT[30] model adopts unsupervised learning mode, and its internal structure is shown in Figure 1. In the training process, GPT still uses the traditional one-way language model training method, but uses Transformer network to replace the LSTM network. The pre-training stage is mainly to train the model parameters on a large-scale unlabeled corpus, and then fine-tune the model according to specific tasks to obtain better portability.

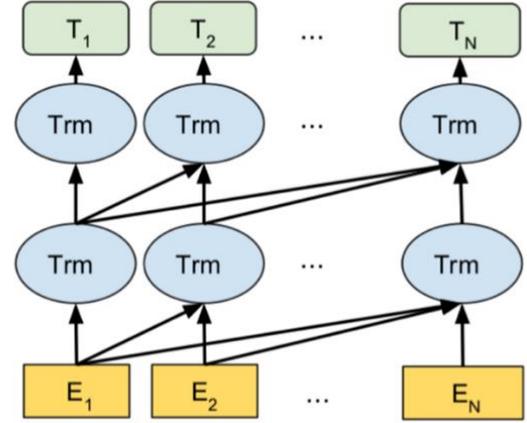

Fig. 1. GPT model structure diagram.

The training regimen for the GPT model incorporates a multi-layer Transformer network along with a softmax layer. The precise computational process is outlined below:

1. First, the final input word vector ($h_0$) is obtained through linear calculation, as shown in the following equation:

$$h_0 = e_W + p_W \quad (4)$$

2. Then, the input word vector ($h_0$) is transferred to the multi-layer Transformer network for processing. The calculation process is shown in equation:

$$h_i^{(l)} = \text{TransformerBlock}(h_{i-1}^{(l)}, n) \quad (5)$$

3. Finally, the prediction probability of the label is obtained through softmax layer. The calculation process is shown in equation below:

$$P(y | x) = \text{softmax}(h_W) \quad (6)$$

When dealing with specific tasks, the task-related data set (*D*) is first passed to the pre-trained language model through the input layer for fine-tuning operations. The calculation process is as follows:

1. Input sequence representation $(x_1, x_2, \ldots, x_m)$ for each sample passed into softmax layer and calculated the prediction probability of the label, as shown in equation:

$$P(y | x) = \text{softmax}(h_{W_1}) \quad (7)$$

2. Compute the cross-entropy loss function based on the predicted probability and actual label, as depicted in the equation below:

$$L(\mathbf{D}) = -\sum_{i=1}^{m} \log(P_{y_i}(x_i)) \qquad (8)$$

In the GPT model, the term ($e_W$) denotes the word vector associated with the text's words, the final input word vector ($h_0$) can be obtained through linear calculation. When dealing with a specific task, the task-related data set ($D$) is first passed to the pre-trained language model through the input layer for fine-tuning operations, and then the prediction probability of the label is obtained through the fully connected layer.

Although the GPT model introduces new ideas in the field of NLP, the structure of its one-way Transformer encoder limits the model's ability to obtain complete information when it comes to understanding text context[31]. Consequently, to address this shortcoming, Google introduced the BERT (Bidirectional Encoder Representations from Transformers) model in 2018, employing a bidirectional Transformer encoder to facilitate comprehensive two-way text understanding.

The design of BERT[32] model greatly expands the representational ability of language models. By utilizing bidirectional encoders, BERT can simultaneously capture contextual information at every position in the text sequence, resulting in a better understanding of the meaning and context of words in a sentence. This bidirectional nature makes BERT more accurate and comprehensive in terms of word representation and semantic understanding.

The architectural framework of the BERT Model is depicted in Figure 2. Throughout the training phase of BERT, two novel tasks are integrated: the Masked Language Model (MLM). This approach, by randomly masking certain words and requiring the model to predict the obscured terms, compels the model to utilize contextual information for comprehensive sentence understanding.

The next sentence prediction task involves assessing the correlation between two successive sentences within a provided text. This task aids the model in gaining a deeper comprehension of the logical and semantic connections existing between sentences.

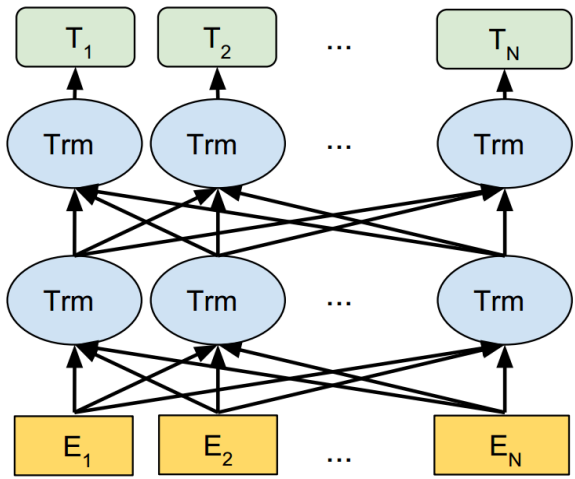

Fig. 2. BERT model structure diagram.

Through these innovative training methods, BERT models can more fully understand the relationship between semantics and sentences in text, thus improving their performance in various NLP tasks. Its powerful semantic representation and portability make BERT one of the important milestones in the current NLP field, bringing breakthrough progress for many natural language processing tasks.

Although BERT pre-training model has advantages in bidirectional modeling of training corpus, the mask language model (MLM) training method it uses has some defects. In practice, BERT models do not use the "[MASK]" mechanism during the fine-tuning process, which leads to a difference between the pre-training phase and the fine-tuning phase. In addition, the simultaneous operation of "[MASK]" on multiple words in the training corpus also restricts the model from learning semantic associations between words.

To address the challenges posed by BERT and mitigate the constraint of traditional autoregressive language models limited to unidirectional modeling, the researchers proposed the XLNet pre-trained language model. Based on the Transformer architecture, XLNet introduces the Permutation Language Model (PLM) mechanism, which aims to solve the shortcomings of BERT. The model calculation process of XLNet is shown in Figure 3.

XLNet[33] model still uses one-way autoregressive model for feature extraction of text words, but it is different from the traditional autoregressive model in terms of input data. Previously, the model obtained the correct order of the words in the sentence, but now it obtains the random permutations of the words in the original sentence. By training fully arranged text statements, XLNet can capture bidirectional contextual feature sequences of the training corpus. To further improve the model's performance when dealing with long text, XLNet also introduced Transformer-XL and relative position coding. Through these innovations, XLNet model not only overcomes the shortcomings of BERT model, but also improves the modeling ability of text semantic features, thus achieving better performance in natural language processing tasks.

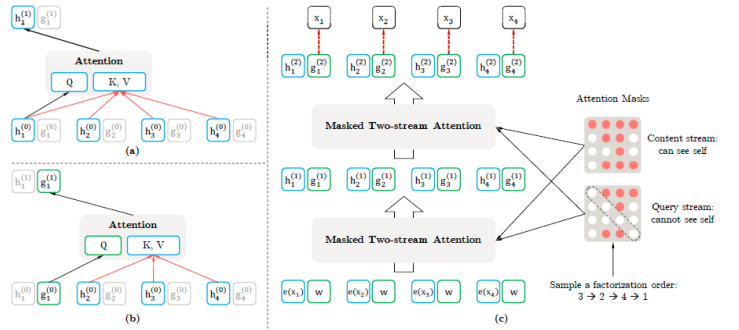

Fig. 3. Pyramid pooling structure.

*C. Kknowledge graph*

The construction of knowledge graph [34] is a complex and continuous process, in which a variety of methods and techniques are applied. Currently, construction methods broadly fall into two categories: top-down and bottom-up approaches. The top-down approach firstly summarizes the ontology structure of knowledge graph in general, and then extracts entities and relationships according to the ontology. In the bottom-up approach, entities and relationships are extracted from real world data, and then the ontology structure of knowledge graph is gradually summarized. In the medical field, the bottom-up approach is usually adopted because of the complexity and diversity of knowledge in the medical field, and the bottom-up approach is more able to make full use of existing medical data while adapting to the constant changes and developments in the medical field.

As shown in Figure 4, the process of building a knowledge graph from the bottom up generally includes the following main stages:

The initial stage involves knowledge extraction, which constitutes the core process of constructing a knowledge graph. This phase primarily involves automatically extracting structured information from semi-structured and unstructured data, encompassing the identification and extraction of entities, relationships, and entity attributes. In the field of medicine, this may involve extracting medical entities (such as diseases, drugs, symptoms, etc.) and relationships between entities from multiple sources such as textual data, medical literature, medical records, etc.

The second stage is knowledge fusion [35], the objective is to consolidate and refine the knowledge extracted from various data sources to eliminate redundancies and errors, thereby enhancing knowledge quality. Knowledge fusion typically involves two main components: entity linking and knowledge merging. Entity linking is the process of correctly linking the identified entity objects in the free text to the target entity objects in the knowledge base, while knowledge merging is to obtain knowledge from the knowledge base constructed by the third party or the existing structured data to update and supplement the current knowledge graph.

The final phase is the knowledge processing stage, primarily aimed at further refining and processing the knowledge that has been extracted and integrated. The ontology construction phase encompasses the calculation of entity parallelism similarity, the extraction of hierarchical relationships among entities, and the generation of ontologies. Knowledge reasoning then delves deeper into uncovering latent knowledge within entities or rectifying incorrect relationships. Quality assessment measures the credibility of knowledge through human participation or using mathematical models to ensure the integrity.

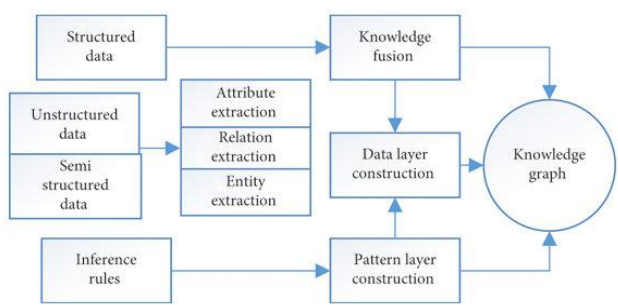

Fig. 4. Knowledge graph construction process.

## IV. MEDICAL TEXT MINING MODEL BASED ON ATTENTION MECHANISM-KNOWLEDGE GRAPH

In this paper, a text mining model combining Attention mechanism and Medical Knowledge Graph Attention Network (MedKG) is proposed. This model combines attention mechanism and medical knowledge graph, and uses pre-trained language model to represent text features, so as to realize more accurate and effective medical text mining.

The MedKG model is designed to overcome the limitations of traditional medical text mining models in processing large-scale heterogeneous medical data and achieving highly personalized recommendations. First, we use the medical knowledge graph to construct a graph structure of multi-source heterogeneous data, where nodes represent different data entities (such as patients, diseases, genes, etc.), and edges represent relationships or interactions between entities. We then introduced an attention mechanism to automatically capture important information and filter out irrelevant information.

In the MedKG model, language model to represent the features of medical texts. Such a pre-trained model can learn rich semantic information from large-scale medical corpus, so as to better understand the meaning of medical text. By combining attention mechanism and medical knowledge graph, we can effectively integrate multi-source heterogeneous data such as clinical characteristics, genetic information and lifestyle habits of patients to build a more accurate representation model of patient health status.

At the heart of the MedKG model is the design of a novel multi-scale fusion mechanism that dynamically adjusts attention allocation strategies to enable highly customized analysis of individual cases. By taking into account the contribution of different characteristics to predictive goals (such as disease diagnosis, treatment choice, etc.), our model can more accurately provide personalized medical advice to patients.

The operation steps of the MedKG model can be summarized in the following key steps: First, the model receives raw medical text data as input, which may include patient clinical records, disease descriptions, medical papers, etc. These text data undergo pre-processing and feature extraction steps for subsequent processing and analysis. Then, the MedKG model uses the pre-trained language model for text representation learning. Pre-trained language models are able to learn semantic information from large-scale text corpora and map each word or subword to a representation in a high-dimensional vector space. The MedKG model then combines text data with a medical knowledge graph by experts in the field of medicine and contains various medical entities (such as diseases, drugs, symptoms, etc.) and the relationships between them. The model uses this knowledge graph to enrich the semantic information of text data and construct a graph structure of heterogeneous data from multiple sources. The MedKG model then applies attention mechanisms to dynamically capture important information in textual data and knowledge graphs. The attention mechanism can automatically learn the degree of correlation between different entities in the text and knowledge graph, thus strengthening the representation of important information, filtering out irrelevant information, and improving the performance of the model. Finally, through the designed multi-scale fusion mechanism, the model integrates and summarizes the information extracted from the text data and knowledge graph. This model can be used for personalized medical advice and prediction to provide patients with more accurate medical services. In summary, MedKG model realizes deep mining and personalized analysis of medical texts through the combination of pre-trained language model, medical knowledge graph and attention mechanism, and provides a new solution for medical decision-making.

In summary, MedKG model realizes deep mining and personalized analysis of medical texts through the combination of pre-trained language model, medical knowledge graph and attention mechanism, and provides a new solution for medical decision-making. MedKG model combines attention mechanism, medical knowledge graph and pre-trained language model to provide a new solution for medical text mining task, which has important theoretical and practical significance.

## V. Experimental analysis

### A. Data set

In our experiment, we chose a medical text dataset named "MedicalRecords" as the basis for verification. The dataset is derived from the hospital's electronic medical record system and contains the clinical records of 1,000 patients. These records include diagnoses, treatment protocols, medication use, surgical records, and laboratory test results. These information are presented in text form, providing a wealth of medical information and a sufficient data base for our research.

Before conducting the experiment, we pre-processed the data set in detail. First, we cleaned the text, removing special characters, punctuation, and meaningless whitespace to keep the data clean and consistent. Then, we used a word segmentation tool to divide the text into sequences of words or sub-words for further processing and analysis. On this basis, we carried out stop word filtering to remove common stop words. In addition, we also performed stem extraction or morphological merging of words to convert words to their basic form, thus reducing the size of the vocabulary and the complexity of the model. In the final stage of pre-processing, we built a medical knowledge graph based on the text content, including entities (such as diseases, drugs, symptoms, etc.) and the relationships between them to enrich the semantic information of the text data. This approach is supported by the methodological insights provided by Li et al.[36], who demonstrated the efficacy of constructing accessibility linked data from publicly available datasets, a process that parallels our creation of a medical knowledge graph to enhance the semantic depth of our dataset for more effective machine learning applications. At the same time, we generated corresponding labels based on the diagnosis results and treatment plans in the clinical records for supervised learning of the model

Through the above pre-processing steps, we preprocess the original medical text data to a format that is suitable for model processing., and provide the model with rich semantic information and labels for experimental verification and performance evaluation. The execution of these steps not only ensures the quality and consistency of the data, but also provides the model with sufficient information.

### B. Evaluation indicators

The classification task evaluation index used in this paper is calculated based on confusion matrix[37]. The confusion matrix divides the sample into four categories: Correct Positives (CP/TP), Correct Negatives (CN/TN), Misclassified Positives (MP/FP), and Missed Positives (NP/FN) – these metrics play a pivotal role in assessing and refining the performance of the Government Information Hybrid Classification Model (GIHCM). CP indicates the count of positively-labeled instances correctly identified, CN reflects the accurate classification of negatively-labeled instances, MP signifies the number of negatives misjudged as positives, and NP represents the positive samples mistakenly categorized as negatives. Collectively, these metrics construct the confusion matrix for the classification task, offering insight into the model's predictive accuracy across different classes. Thorough analysis of the confusion matrix enables a more holistic evaluation of the classification model's capabilities, with the corresponding category classifications detailed in Table 1.

TABLE I. CONFUSION MATRIX OF SAMPLES

|  | positive sample | negative sample |
|---|---|---|
| positive | TP | FN |
| negative | FP | TN |

When gauging model proficiency, evaluative metrics like Accuracy and F1 Score frequently serve as barometers for measuring model competence. Accuracy, essentially, measures the ratio of accurately predicted instances relative to the entire sample population, serving as a broadly accepted benchmark in evaluating classification models' performance. Its computation typically adheres to the following formula:

$$\text{Accuracy} = \frac{TP + TN}{TP + TN + FP + FN} \quad (9)$$

The F1 Score represents a harmonized metric that amalgamates Precision and Recall, capturing both the accuracy and comprehensiveness of the model. This metric is derived using the harmonic mean, according to the formula presented below:

$$F1 = \frac{2 \times \text{Precision} \times \text{Recall}}{\text{Precision} + \text{Recall}} \quad (10)$$

Precision measures the proportion of cases correctly identified as positive by the model out of all labeled as positive, while Recall assesses how well the model identifies actual positives within the dataset. The F1 Score, ranging from 0 to 1, with values closer to 1 indicating superior model efficacy, provides a comprehensive view of model performance. These dual metrics facilitate a thorough evaluation of the model's classification capabilities. Accuracy gauges the fraction of all predictions made by the model that are accurate, whereas the F1 Score considers both precision and recall, offering a more nuanced reflection of the model's effectiveness across various categories.

### C. Experimental setup

In the experimental part of this study, we divided the data set "MedicalRecords" by 8:1:1 and constructed the data configuration including training set, tuning and performance evaluation. For the parameter configuration of the MedKG model, we carefully design the following scheme: Adopt 4-layer Transformer network architecture, configure 8 attention heads in each layer, and set the hidden unit dimension to 256, aiming at capturing complex semantic relationships while maintaining computational efficiency. The learning rate is set to 0.0005 to balance the convergence rate; Set the batch size to 64 to optimize training efficiency and model stability. In terms of training strategy, we use Adam optimizer and cross entropy loss function to drive model learning, and implement learning rate attenuation strategy to dynamically adjust learning rate to promote fine adjustment in the late training period. Moreover, an early stopping mechanism was integrated as a strategy to mitigate overfitting. This involved halting the training process when there was no substantial improvement in the loss on the validation set, thereby safeguarding the model's ability to generalize. By meticulously defining parameters and employing various strategies, we methodically assessed the MedKG model's efficacy in medical text mining tasks, thus ensuring the credibility and validity of our experimental findings.

*D. Experimental result*

In Table 2, we present experimental results under different baseline models, including MedKG, GPT, BERT, and XLNet. The models were compared for performance in medical text mining tasks, evaluated on measures including accuracy (ACC) and F1 scores. Under our proposed MedKG model, the accuracy rate (ACC) reached 96.32% and the F1 score reached 95.24%. By combining attention mechanism and medical knowledge graph, MedKG model can realize deep mining of medical texts, and can effectively capture the associated information among medical entities, thus improving the performance of text classification tasks.

The GPT model achieved an accuracy of 89.21%, with the F1 score marginally higher at 93.12%. GPT, an unsupervised learning language model, is effective in medical text mining tasks, though it exhibits a slight performance decline relative to the MedKG model. The BERT model recorded an accuracy of 90.97% and an F1 score of 90.28%. As a pre-trained language model utilizing a bidirectional Transformer encoder, BERT excels in natural language processing tasks but performs slightly less effectively than MedKG in medical text mining. The XLNet model demonstrated an accuracy of 92.68% and an F1 score of 89.36%. XLNet, which introduces a permutation language model (PLM), shows a performance slightly superior to BERT in experiments, yet it still falls short of the MedKG model.

To sum up, MedKnowGraph model achieves the best performance in medical text mining tasks. It can better mine semantic information in text by combining attention mechanism and medical knowledge graph, thus improving the accuracy and efficiency of text classification.

TABLE II. EXPERIMENTAL RESULTS UNDER DIFFERENT SETS

| MODEL | ACC | F1 |
| --- | --- | --- |
| MedKG | 96.32 | 95.24 |
| GPT | 89.21 | 93.12 |
| BERT | 90.97 | 90.28 |
| XLNet | 92.68 | 89.36 |

*E. Ablation experiment*

Table 3 presents a comparison of ablation results between two models, MedKG and MedG. The experiments aimed to evaluate the effect of critical components incorporated into the MedKG model on its performance. In the MedKG model, an attention mechanism and a medical knowledge graph are employed to enhance the deep mining of medical texts. The experimental results indicate that the MedKG model attains an accuracy (ACC) of 96.32% and an F1 score of 95.24%. This demonstrates the model's ability to effectively leverage the attention mechanism and medical knowledge graph to capture pertinent information among medical entities, thus enhancing text classification performance. Conversely, the MedG model, which excludes the attention mechanism and medical knowledge graph, relying solely on traditional text mining methods, yields an accuracy of 92.61% and an F1 score of 90.89%. This marginal decrease in performance compared to the MedKG model underscores the beneficial impact of incorporating the attention mechanism and medical knowledge graph, which significantly enhance model efficacy in medical text mining tasks. In conclusion, the ablation analysis underscores the critical role played by the attention mechanism and medical knowledge graph in bolstering model performance, thereby enhancing the accuracy and efficiency of text classification tasks.

TABLE III. ABLATION RESULTS

| MODEL | ACC | F1 |
| --- | --- | --- |
| MedKG | 96.32 | 95.24 |
| MedG | 92.61 | 90.89 |

VI. CONCLUSION

In this groundbreaking work, we successfully constructed and validated the MedKG model, an innovative medical text mining framework that creatively combines advanced attention mechanisms with rich medical knowledge graph resources to bring significant performance improvements to medical text classification tasks. Through rigorous empirical analysis, the MedKG model has demonstrated superior performance over traditional approaches and several cutting-edge baseline models such as GPT, BERT, and XLNet on a range of key performance indicators, including accuracy and F1 scores. The acquisition of this significant advantage not only validates the prospectivity and effectiveness of our model design but also further highlights the innovative value of the integration of attention mechanism and medical knowledge graph, which together promote the model to understand and accurately classify the deep semantic meaning of medical texts. The in-depth investigation of the ablation experiment systematically revealed the indispensability of each core component in the MedKG model. Specifically, the decline in performance of the model variant MedG without the support of the attention mechanism or the medical knowledge graph clearly confirms the critical role of these two mechanisms in enhancing the model's performance and ensuring high accuracy and efficient processing power. This finding not only strengthens the theoretical basis of our model design but also provides valuable insights for future research. In addition, the MedKG model's portability and generalization ability evaluation further broadens the boundaries of its application. Experiments show that this model can maintain its excellent performance in any specific medical field, which shows its strong potential and practical application value as a general medical text mining tool. This has far-reaching significance for promoting cross-domain knowledge sharing and improving the intelligence level of the medical decision support system.

In summary, the MedKG model not only represents a major advance in the field of medical text mining, its outstanding contribution to improving the accuracy and efficiency of text classification tasks but also opens up entirely new possibilities for the medical industry. Looking to the future, the MedKG model is expected to become a sharp edge in the hands of medical practitioners, providing more accurate and efficient auxiliary tools for clinical decision-making, thereby promoting the modernization of medical practice, and helping to achieve more personalized and accurate medical services.